\documentclass{article}

\usepackage[final,nonatbib]{neurips_2021}
\usepackage{comment}
\usepackage{enumitem}

\usepackage[utf8]{inputenc} 
\usepackage[T1]{fontenc}    
\usepackage{hyperref}       
\usepackage{url}            
\usepackage{booktabs}       
\usepackage{amsfonts}       
\usepackage{nicefrac}       
\usepackage{microtype}      
\usepackage{xcolor}         

\usepackage[sortcites]{biblatex}
\title{What can Data-Centric AI Learn from\\Data and ML Engineering?}

\author{%
  Neoklis Polyzotis \\
  Databricks
  \And
  Matei Zaharia \\
  Databricks and Stanford
}

\begin{document}

\maketitle

\begin{abstract}
  Data-centric AI is a new and exciting research topic in the AI community, but many organizations already build and maintain various ``data-centric'' applications whose goal is to produce high quality data. These range from traditional business data processing applications (e.g., ``how much should we charge each of our customers this month?'') to production ML systems such as recommendation engines. The fields of data and ML engineering have arisen in recent years to manage these applications, and both include many interesting novel tools and processes. In this paper, we discuss several lessons from data and ML engineering that could be interesting to apply in data-centric AI, based on our experience building data and ML platforms that serve thousands of applications at a range of organizations.
\end{abstract}

\section{Introduction}

Data-centric AI (DCAI) is an exciting new research field that studies the problem of constructing high-quality datasets for machine learning. Although many of the specific challenges in DCAI are new, we can also look at DCAI through the lens of data-centric applications in general. Today, a wide range of important computer applications are ``data-centric,'' in that their main goal is to produce high-quality datasets. These range from traditional business applications (e.g., processing usage records for an electric company to compute how much to charge each customer) to production ML deployments, where the field of ML engineering~\cite{ml-engineering} and new software such as ML platforms~\cite{tfx,michelangelo,mlflow} have arisen to reliably process new training data, update models, and monitor performance. Both data engineering and ML engineering are evolving rapidly, with a wide range of new tools and processes being adopted that have made data-centric applications easier to build~\cite{tfx,tfdv,dbt,great-expectations,michelangelo,kubeflow,mlflow}.

In this paper, we discuss how several lessons from data and ML engineering may apply to data-centric AI, based on the authors' experience working on open source and commercial platforms such as MLflow~\cite{mlflow} and TFX~\cite{tfx}.
We view ``data-centric AI'' as the problem of designing data collection, labeling, and quality monitoring processes for datasets to be used in machine learning.
Although many aspects of DCAI are different from traditional data applications (e.g., figuring out how to interact with human annotators), others are similar, and some lessons about the challenges in other data-centric applications may motivate new research in DCAI.

We organize the paper around five lessons we found interesting in production systems:
\begin{enumerate}
    \item Data and AI applications need to run and train continuously, not just once.
    \item Production deployment workflows are often code-centric, not model- or data-centric.
    \item Data monitoring must be actionable.
    \item End-to-end support for versioning code and data is immensely helpful.
    \item Some applications are not allowed to show data to human annotators or developers.
\end{enumerate}

\section{Data and AI Applications Must be Updated Continuously}

Much of the current research discussion around data-centric AI focuses on how to create a high-quality dataset for a particular ML problem \emph{once}.
For instance, a research team may carefully collect diverse training images, define classes, select an annotation strategy (and a way of combining labels from multiple annotators), label and evaluate data to create a high-quality, ImageNet-like dataset for their domain: diagnosing patients from CAT scans, identifying defects in an assembly line, etc.
While this type of effort is definitely valuable, we observed that for most production AI and data applications, \emph{the datasets and models need to be updated continuously}, and these continuous updates are the most challenging aspect for development teams.
For example, the CAT-based diagnostic tool may need new data and retraining every month because new CAT machines have been deployed at hospitals, or the existing machines received a software update, or errors were found with the previous model; the assembly line monitor may need to support a new product being assembled, or changes in the materials or sensors used; etc.
Even applications that seem to target a ``static'' problem often need to be updated due to changes in the way the data is gathered (e.g., new CAT software even though the underlying medical conditions are the same) or changing problem definitions (e.g., identifying more diseases).
Applications that target more dynamic problems, such as blocking illegal content on a social network or fraud detection, may need to be updated hourly or even faster.

This continuous execution model significantly changes the problems that DCAI needs to consider. Instead of selecting a diverse training dataset once, real-world users probably need a tool that will automatically create new datasets each day. Instead of defining classes once, users may want algorithms that can handle an evolving taxonomy and use old data for some classes together with newer data for others. The same is true for labeling: DCAI tools should ideally be prepared to handle changes in label data stemming from changes to the annotation UI, the pool of annotators, or the problem definition. Within current DCAI research, some approaches, such as weak supervision via labeling functions~\cite{data-prog}, are well-poised to handle changes in the task definition (one can simply run new labeling functions over old data), while many approaches relying on annotators are not. In addition, continuous execution significantly changes human developers' workflows. For example, while ML researchers may be able to review a few training examples and metrics manually when building their ``ImageNet'' dataset once, they probably do not want to do this every day, so the continuous data collection, label requisition and retraining process likely needs to be fully automated, with automated alerts when things appear to be going wrong. In our experience, most production ML and data engineering teams fully automate their workflows and monitor them through alerts in this manner using the many tools in this space~\cite{mlflow,tfx,great-expectations}, even for highly critical applications. Doing this for the various phases of data-centric AI is an interesting challenge.

\section{Production Deployment Workflows are Often Code-Centric}

One possibly surprising fact we learned from many teams who deploy ML is that the training code, and not the model, defines the boundary between experimentation and production. Specifically, a typical viewpoint of ML encompasses three major artifacts: the input dataset, the training code, and the resulting trained model. The latter is a central artifact, since it represents the final result of the ML process and can also define the boundary between experimentation and production: during experimentation, a data scientist iteratively develops a high-quality model, then she hands it off to an ML engineer who is only responsible for deploying it in production. 

Many production teams eschew this viewpoint and instead consider the training code as the major artifact at the boundary between experimentation and production. Much in the same way that a software engineer will compile the code of a program against a specific build environment and then run the resulting binary, ML engineers want to ``compile'' the training code against a ML build environment that includes both code and data dependencies, and then run the resulting model in production.
This approach is partly due to the continuous execution challenge we discussed earlier (the ML engineer will need to retrain the model periodically with new data), but also due to the need for reproducibility, testability, and ensuring that the model is compatible with production parameters that are outside the data scientist's control. To illustrate incompatibilities that can arise between experimentation and production and can be caught through this ``compilation'' process, the data scientist may try to use features not yet available in the serving environment, or use training operations that are not yet supported in the serving stack. Blindly deploying a model to production can create many hard-to-diagnose problems at runtime.

This notion of code-centric ML deployment creates intriguing challenges for data-centric AI.
First, human annotation is often significantly more expensive than model training, so it seems untenable to require separate ``development'' and ``production'' annotation processes.
Teams will probably instead require ways to iterate on experimental and production annotation code separately while reusing labels when possible. This requires careful tracking of which version of the UI was used to produce each annotation, and perhaps special care to not show the same examples to the same annotators in different UIs. Second, much like conventional production systems, the ``build environment'' for an ML application evolves over time to include new dependencies, which in this case correspond to new datasets or existing datasets with changed semantics. An example of the latter is changing the set of classes labelers can provide. These operations are common and are already facilitated by infrastructure such as Feature Stores. Note that each data evolution essentially results in a new version of the build environment, without necessarily replacing or deprecating previous versions. In other words, we end up with multiple build environments (each with its own data and system dependencies) that co-exist in the same production system. Furthermore, within each environment, there may be continuous or periodic updates to its datasets --- for example, we may still want to annotate data using an old version of the annotation UI while testing a new one that produces annotations in a different format. Overall, we are no longer talking about a single dataset that is used for ML (which corresponds to the typical dataset-code-model view of ML), but about a ``multiverse'' of datasets, where each universe corresponds to the evolving datasets of a specific version of the build environment. Enabling these many versions of an ML application to share annotations and labeling resources efficiently and correctly is an interesting problem.

\section{Data Monitoring Must Be Actionable}

Given the dynamicity of the ML applications, production teams rely heavily on data monitoring to track changes in new training data, serving data, and production models' outputs. 
For instance, we may want to verify that there are not too many missing values in the data, or that feature values come from predefined domains.
We may also monitor the fit of the data for specific ML tasks, e.g., whether the training data provides enough coverage for existing labels.
It is worth noting that one of the most effective monitoring tools in practice is simply the data schema: providing an expected structure for the data, including ranges for each field, gives users easy-to-understand messages when data is out of range and an easy way to change the accepted schema.
This approach is standard in databases and increasingly adopted for ML~\cite{tfdv, great-expectations, deequ}.

We observed that one critical factor in the success of monitoring tools is whether their outputs are \emph{actionable}: that is, can users quickly diagnose the problem that caused an alert, or reconfigure the monitoring system to prevent false alarms if the alert was erroneous.
Without actionable outputs, engineers will often silence and ignore the alerts, thus essentially turning off data monitoring~\cite{tfdv}. One approach is to focus on alerts that are informative. For example, consider two versions of a distribution-shift alert: ``KL divergence between serving and training data has exceed pre-specified threshold $x$'' vs ``the genre feature seems to have shifted in distribution, with horror being the most frequent value (was: comedy)''. The second alert is more specific and informative, even if it less powerful in terms of capturing more general shifts. Another approach is to tie data monitoring to the semantics of downstream training, so that it becomes possible to triage errors and alerts based on their expected effect on model quality. For instance, if data monitoring is aware of the data-to-models lineage and associated metrics of feature importance, then the previous example alert on distribution shift can also mention whether ``genre'' is an important feature. 
At the extreme, the monitoring system can suppress alerts for unimportant features.
Overall, DCAI tools should also aim for actionable outputs when designing methods to manage data quality.

\section{Platforms Should Support End-to-End Versioning}

Apart from actionable data monitoring, the biggest trend we have seen in data engineering and ML platforms recently is end-to-end support for versioning data, code, and derived artifacts such as ML models.
Versioning code is already a software engineering best practice, but today, data management systems ranging from SQL data warehouses to file-oriented data lakes have also added support for versioning: each time users update data, the system retains copies of the old data, and makes it easy to roll back to them~\cite{delta-lake,lakefs,dvc}. For machine learning specifically, platforms such as MLflow, DVC, Michelangelo and TFX also support versioning models and tracking the data versions they used~\cite{mlflow,dvc,michelangelo,tfx}. 
Many of these systems also support cheap copy-on-write ``branching'' from an old version of a dataset if a user wants to explore changes based on that version.
In our experience, users uniformly appreciate these versioning features and use them to debug, test, and improve their workflows.

Within data-centric AI, first-class support for versioning would mean that DCAI systems need to track multiple versions of the UIs and tasks shown to human annotators, track which version each response came from, and reason about how to combine different datasets and annotations. It may help to build DCAI systems on top of existing versioned data stores, but the DCAI systems will likely also need to take special care in how they send tasks to annotators if they wish to avoid giving the same annotator multiple versions of the same task. Implementing DCAI workflows on top of versioned code, data and model management systems may also greatly improve reproducibility and help downstream users establish trust in datasets.

\section{Some Applications are Not Allowed to Show Data to Humans}

Another common viewpoint in research is that the ML engineers and human annotators have access to the training data and inference data of a model. After all, it is standard to examine the training data in order to develop better models or to inspect the inference data in order to debug issues.

Again, there are important production ML settings that break this assumption. For example, consider a model that auto-completes text in a messaging application. Any access to the training or inference data can leak sensitive information about individuals, and so production teams opt to lock down access only to principals corresponding to production jobs (for training and inference) and to disallow any access from human operators. Another example is models deployed on customers' private infrastructure, e.g., a camera system in a secure facility that never sends data back to the ML vendor.

This setting poses several challenges, since humans need to understand and somehow annotate the underlying data in order to develop effective training methods or debug issues with quality. Thankfully, techniques from data engineering can provide some relief here as well.
For ML model designers, one common approach is to allow access on highly aggregated data views so that the human operator can understand the ``shape'' of the data without inspecting individual data points.
Differentially-private query processing techniques~\cite{pinq, dp-viz} are also promising here, as they allow data scientists some freedom to explore the data, while bounding the risk of leaking sensitive information.
These approaches can help designers create programmatic weak supervision rules to help label the data, check how much of the dataset is covered by them, evaluate their model's quality on slices, or even generate model cards~\cite{model-cards}.
Another approach, which is useful for leveraging human annotators, is to ``simulate'' the data, i.e., to generate a proxy synthetic dataset that the human operator can use in order to optimize training parameters or to debug model problems~\cite{synthetic-data-net,synthetic-data-med, dp-data-release}. 
A third approach, analogous to unit testing, is to manually prepare a small training dataset that covers common scenarios, but this can be challenging in continuously changing domains, such as an auto-complete function that keeps up with the latest celebrity names and song titles.

Teams may also opt for a different approach where \emph{some} data access is permitted for specific reasons (e.g., to debug model errors in production if a customer gives the team permission), but all such accesses are logged (ideally in an automated fashion) for the purpose of auditing. 
This approach raises ML challenges about how to combine the small amounts of operator-visible data with other forms of supervision and model evaluation to produce a model that reliably fixes the reported problem.

\section{Conclusion}

Data-centric AI tackles the key problem of improving data quality to improve ML, but making DCAI effective in production applications may require solving new challenges. In our experience, production data and ML applications need to be updated continuously to handle changing conditions, without having their designers in the loop; they often need to fit into a software engineering process that centers around code; and they may even need to run without ever showing production data to humans. Fortunately, there are also ideas from data and ML engineering, such as end-to-end version tracking and actionable monitoring, that could be extended to help DCAI tackle these problems.

\printbibliography

@article{ml-engineering,
  author    = {Konstantinos Katsiapis and
               Abhijit Karmarkar and
               Ahmet Altay and
               Aleksandr Zaks and
               Neoklis Polyzotis and
               Anusha Ramesh and
               Ben Mathes and
               Gautam Vasudevan and
               Irene Giannoumis and
               Jarek Wilkiewicz and
               Jiri Simsa and
               Justin Hong and
               Mitchell Trott and
               No{\'{e}} Lutz and
               Pavel A. Dournov and
               Robert Crowe and
               Sarah Sirajuddin and
               Tris Brian Warkentin and
               Zhitao Li},
  title     = {Towards {ML} Engineering: {A} Brief History Of TensorFlow Extended
               {(TFX)}},
  journal   = {CoRR},
  volume    = {abs/2010.02013},
  year      = {2020},
  url       = {https://arxiv.org/abs/2010.02013},
  eprinttype = {arXiv},
  eprint    = {2010.02013},
  bibsource = {dblp computer science bibliography, https://dblp.org}
}

@inproceedings{tfdv,
  author    = {Eric Breck and
               Neoklis Polyzotis and
               Sudip Roy and
               Steven Whang and
               Martin Zinkevich},
  title     = {Data Validation for Machine Learning},
  booktitle = {Proceedings of Machine Learning and Systems 2019, MLSys 2019, Stanford,
               CA, USA},
  publisher = {mlsys.org},
  year      = {2019},
}

@inproceedings{tfx,
  author    = {Denis Baylor and
               Eric Breck and
               Heng{-}Tze Cheng and
               Noah Fiedel and
               Chuan Yu Foo and
               Zakaria Haque and
               Salem Haykal and
               Mustafa Ispir and
               Vihan Jain and
               Levent Koc and
               Chiu Yuen Koo and
               Lukasz Lew and
               Clemens Mewald and
               Akshay Naresh Modi and
               Neoklis Polyzotis and
               Sukriti Ramesh and
               Sudip Roy and
               Steven Euijong Whang and
               Martin Wicke and
               Jarek Wilkiewicz and
               Xin Zhang and
               Martin Zinkevich},
  title     = {{TFX:} {A} TensorFlow-Based Production-Scale Machine Learning Platform},
  booktitle = {Proceedings of the 23rd {ACM} {SIGKDD} International Conference on
               Knowledge Discovery and Data Mining, Halifax, NS, Canada, August 13
               - 17, 2017},
  pages     = {1387--1395},
  publisher = {{ACM}},
  year      = {2017},
}

@article{deequ,
  author    = {Sebastian Schelter and
               Dustin Lange and
               Philipp Schmidt and
               Meltem Celikel and
               Felix Bie{\ss}mann and
               Andreas Grafberger},
  title     = {Automating Large-Scale Data Quality Verification},
  journal   = {Proc. {VLDB} Endow.},
  volume    = {11},
  number    = {12},
  pages     = {1781--1794},
  year      = {2018},
  url       = {http://www.vldb.org/pvldb/vol11/p1781-schelter.pdf},
  doi       = {10.14778/3229863.3229867},
  bibsource = {dblp computer science bibliography, https://dblp.org}
}

@online{michelangelo,
  author = {Jeremy Hermann and Mike Del Balso},
  title = {Meet Michelangelo: Uber’s Machine Learning Platform},
  year = 2017,
  url = {https://eng.uber.com/michelangelo-machine-learning-platform/},
}

@online{kubeflow,
    key={kubeflow},
    title={Kubeflow: The Machine Learning Toolkit for Kubernetes},
    url = {https://kubeflow.org}
}

@article{mlflow,
  author    = {Matei Zaharia and
               Andrew Chen and
               Aaron Davidson and
               Ali Ghodsi and
               Sue Ann Hong and
               Andy Konwinski and
               Siddharth Murching and
               Tomas Nykodym and
               Paul Ogilvie and
               Mani Parkhe and
               Fen Xie and
               Corey Zumar},
  title     = {Accelerating the Machine Learning Lifecycle with MLflow},
  journal   = {{IEEE} Data Eng. Bull.},
  volume    = {41},
  number    = {4},
  pages     = {39--45},
  year      = {2018},
  bibsource = {dblp computer science bibliography, https://dblp.org}
}

@online{dbt,
    key={dbt},
    title={dbt},
    url = {https://www.getdbt.com}
}

@online{great-expectations,
   key={great-expecations},
   title={Great Expecations},
   url={https://greatexpectations.io}
}

@online{lakefs,
   key={lakefs},
   title={LakeFS: Atomic Versioned Data Lake},
   url={https://lakefs.io}
}

@online{dvc,
   key={dvc},
   title={DVC: Data Version Control},
   url={https://dvc.org}
}

@inproceedings{pinq,
  author    = {Frank McSherry},
  title     = {Privacy integrated queries: an extensible platform for privacy-preserving
               data analysis},
  booktitle = {{SIGMOD 2019}},
  pages     = {19--30},
  publisher = {{ACM}},
  year      = {2009},
}

@article{delta-lake,
author = {Armbrust, Michael and Das, Tathagata and Sun, Liwen and Yavuz, Burak and Zhu, Shixiong and Murthy, Mukul and Torres, Joseph and van Hovell, Herman and Ionescu, Adrian and \L{}uszczak, Alicja and undefinedwitakowski, Micha\l{} and Szafra\'{n}ski, Micha\l{} and Li, Xiao and Ueshin, Takuya and Mokhtar, Mostafa and Boncz, Peter and Ghodsi, Ali and Paranjpye, Sameer and Senster, Pieter and Xin, Reynold and Zaharia, Matei},
title = {Delta Lake: High-Performance ACID Table Storage over Cloud Object Stores},
year = {2020},
issue_date = {August 2020},
publisher = {VLDB Endowment},
volume = {13},
number = {12},
journal = {Proc. VLDB Endow.},
month = aug,
pages = {3411–3424},
numpages = {14}
}

@inproceedings{synthetic-data-net,
author = {Lin, Zinan and Jain, Alankar and Wang, Chen and Fanti, Giulia and Sekar, Vyas},
title = {Using GANs for Sharing Networked Time Series Data: Challenges, Initial Promise, and Open Questions},
year = {2020},
publisher = {Association for Computing Machinery},
address = {New York, NY, USA},
booktitle = {Proceedings of the ACM Internet Measurement Conference},
pages = {464–483},
numpages = {20},
location = {Virtual Event, USA},
series = {IMC '20}
}

@article{dp-viz,
  author    = {Dan Zhang and
               Ali Sarvghad and
               Gerome Miklau},
  title     = {Investigating Visual Analysis of Differentially Private Data},
  journal   = {{IEEE} Trans. Vis. Comput. Graph.},
  volume    = {27},
  number    = {2},
  pages     = {1786--1796},
  year      = {2021},
  
  doi       = {10.1109/TVCG.2020.3030369},
  
  bibsource = {dblp computer science bibliography, https://dblp.org}
}

@article{synthetic-data-med,
	author = {Tucker, Allan and Wang, Zhenchen and Rotalinti, Ylenia and Myles, Puja},
	da = {2020/11/09},
	doi = {10.1038/s41746-020-00353-9},
	isbn = {2398-6352},
	journal = {npj Digital Medicine},
	number = {1},
	pages = {147},
	title = {Generating high-fidelity synthetic patient data for assessing machine learning healthcare software},
	volume = {3},
	year = {2020}
}

@inproceedings{dp-data-release,
  author    = {Moritz Hardt and
               Katrina Ligett and
               Frank McSherry},
  editor    = {Peter L. Bartlett and
               Fernando C. N. Pereira and
               Christopher J. C. Burges and
               L{\'{e}}on Bottou and
               Kilian Q. Weinberger},
  title     = {A Simple and Practical Algorithm for Differentially Private Data Release},
  booktitle = {Advances in Neural Information Processing Systems 25: 26th Annual
               Conference on Neural Information Processing Systems 2012.},
  pages     = {2348--2356},
  year      = {2012},
}

@inproceedings{model-cards,
author = {Mitchell, Margaret and Wu, Simone and Zaldivar, Andrew and Barnes, Parker and Vasserman, Lucy and Hutchinson, Ben and Spitzer, Elena and Raji, Inioluwa Deborah and Gebru, Timnit},
title = {Model Cards for Model Reporting},
year = {2019},
isbn = {9781450361255},
publisher = {Association for Computing Machinery},
address = {New York, NY, USA},
url = {https://doi.org/10.1145/3287560.3287596},
booktitle = {Proceedings of the Conference on Fairness, Accountability, and Transparency},
pages = {220–229},
numpages = {10},
location = {Atlanta, GA, USA},
series = {FAT* '19}
}

@inproceedings{data-prog,
 author = {Ratner, Alexander J and De Sa, Christopher M and Wu, Sen and Selsam, Daniel and R\'{e}, Christopher},
 booktitle = {Advances in Neural Information Processing Systems},
 editor = {D. Lee and M. Sugiyama and U. Luxburg and I. Guyon and R. Garnett},
 pages = {},
 publisher = {Curran Associates, Inc.},
 title = {Data Programming: Creating Large Training Sets, Quickly},
 volume = {29},
 year = {2016}
}

\end{document}